\documentclass[sigconf]{acmart}
\renewcommand\footnotetextcopyrightpermission[1]{}
\setcopyright{none}
\acmISBN{}

\newtheorem*{runningexample}{Running example}

\usepackage{listings}
\usepackage{tabularx}
\usepackage{multirow}
\usepackage{amsmath}
\usepackage{amsfonts}
\usepackage{pifont}
\colorlet{punct}{red!60!black}
\definecolor{background}{HTML}{EEEEEE}
\definecolor{delim}{RGB}{20,105,176}
\colorlet{numb}{magenta!60!black}

\lstdefinelanguage{json}{
    basicstyle=\normalfont\ttfamily,
    numbers=left,
    numberstyle=\scriptsize,
    stepnumber=1,
    numbersep=8pt,
    showstringspaces=false,
    breaklines=true,
    string=[s]{"}{"},
    stringstyle=\color{blue},
    frame=lines,
    backgroundcolor=\color{background},
    literate=
      {:}{{{\color{punct}{:}}}}{1}
      {,}{{{\color{punct}{,}}}}{1}
      {\{}{{{\color{delim}{\{}}}}{1}
      {\}}{{{\color{delim}{\}}}}}{1}
      {[}{{{\color{delim}{[}}}}{1}
      {]}{{{\color{delim}{]}}}}{1},
}
\newcommand{\cmark}{\ding{51}}%
\newcommand{\xmark}{\ding{55}}%
\begin{document}

\title{Using LLMs for Explainable, Data-Driven Insight
Generation from Time Series }

\author{Ria Mundhra}
\email{mundhra.ria@gmail.com}
\affiliation{%
  \institution{University of Oxford}
  \city{Oxford}
  \country{United Kingdom}
}

\author{Gustavo Sato dos Santos}
\affiliation{%
  \institution{Vortexa}
  \city{London}
  \country{United Kingdom}
}
\email{gustavo.santos@vortexa.com}

\author{Michael Benedikt}
\affiliation{%
  \institution{University of Oxford}
  \city{Oxford}
  \country{United Kingdom}
}
\email{michael.benedikt@cs.ox.ac.uk}


\begin{abstract}
Time series forecasts are widely used in decision-critical domains, where they are rarely consumed without accompanying explanations. Producing such explanations is usually a manual and costly process, and attempts to automate it using large language models often suffer from hallucination when applied to temporal data.

We propose a domain-agnostic framework for grounded natural language explanation generation for time series forecasts, illustrated in Figure~\ref{fig:teaser}. The framework consists of three components: (i) extraction of structured explanatory factors from historical analyst-written explanations, (ii) evidence-conditioned explanation generation, and (iii) scalable evaluation for readability, logical consistency, and persuasiveness. The design explicitly constrains generation to verifiable evidence, reducing unsupported claims.

We evaluate the framework on a financial forecasting case study involving the NASDAQ-100 index and a freight pricing case study using data from Vortexa. Results show that generated explanations approached analyst-written explanations in terms of readability, consistency and persuasiveness. These findings demonstrate that grounded explanation generation for time series forecasting can be achieved at scale without domain-specific fine-tuning.
\end{abstract}


\begin{teaserfigure}
  \includegraphics[width=\textwidth]{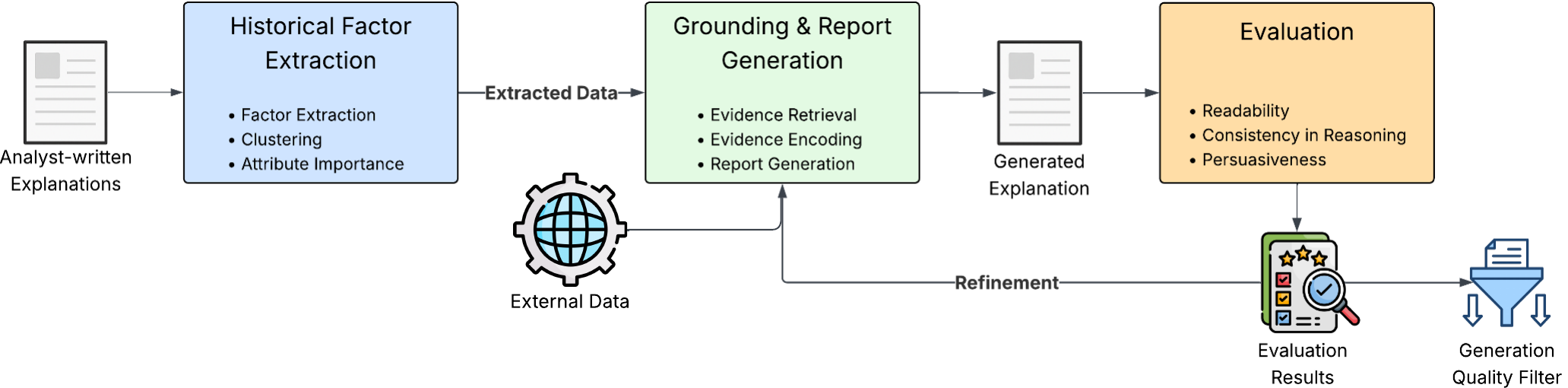}
  \caption{A modular framework for generating grounded natural language explanations for time series forecasts.}
  \label{fig:teaser}
  \Description{Block diagram illustrating the end-to-end framework for grounded explanation generation. On the left, analyst-written explanations are input to a module labelled “Historical Factor Extraction” which includes factor extraction, clustering, and attribute importance. The extracted data, together with external data, flow into a central module labelled “Grounding and Report Generation” consisting of evidence retrieval, evidence encoding, and report generation. This module outputs a generated explanation, which is passed to an “Evaluation” module on the right that assesses readability, consistency in reasoning, and persuasiveness. Evaluation results are shown feeding back into the grounding and report generation module via a refinement loop.}
\end{teaserfigure}

\maketitle

\section{Introduction} \label{sec:intro}
Time series forecasting underpins decision-making in many high-stakes domains \cite{kaushik2020ai, sezer2020financial, brandt2006advances, bechtel2010forecasting}. In such settings, forecasts are rarely consumed in isolation. Instead, stakeholders require explanations that justify predicted trends, contextualize uncertainty, and connect model outputs to observable evidence. These explanations are critical for trust, accountability, and decision-making, particularly when forecasts influence costly or irreversible actions \cite{jiang2025explainable}.

In many such domains, explanatory forecasting remains largely a human-driven process. Domain experts interpret numerical predictions by synthesizing historical patterns, external signals, and domain knowledge into explanations for forecasts. While effective, this approach is expensive, slow, and difficult to scale. As organizations in such settings increasingly demand frequent, on-demand explanations across many assets or systems, the reliance on expert-written narratives becomes a significant bottleneck \cite{park2025analysts}.

Recent advances in large language models (LLMs) suggest a potential path toward automating this process. LLMs can generate fluent, coherent text and integrate heterogeneous information sources, making them attractive candidates for producing explanatory narratives \cite{zhang2023insight, hadi2023large, fei-etal-2024-multimodal}. However, applying LLMs to time series explanation raises fundamental challenges \cite{jiang2025explainable}. Unlike natural language, time series data is numerical, temporally structured, and often exhibits strong autocorrelation, seasonality, and regime shifts. LLMs are not trained to reason directly over such structures and, when used naively, may generate explanations that are factually incorrect, temporally inconsistent, or unsupported by data.

\paragraph{Challenges in LLM Time Series Explanation}
\label{sec:challenges}
The core difficulty lies not in generating text, but in generating \emph{grounded} explanations - explanations in which each claim is explicitly supported by verifiable evidence drawn from the underlying time series data or retrieved external sources provided to the model at generation time. Without appropriate constraints, LLMs may hallucinate causal mechanisms, overemphasize spurious correlations, or introduce narratives that are not justified by the data \cite{jiang2025explainable, merrill2024language}.

Moreover, it is unclear what information should be provided to LLMs to enable faithful explanation generation. Time series forecasting pipelines often expose a large number of potential signals, including raw historical values, derived features, external covariates, and unstructured contextual data such as news reports. Conditioning an LLM on all available information is impractical and may degrade performance by introducing noise. Conversely, overly minimal conditioning risks omitting critical explanatory context necessary for producing meaningful justifications \cite{pang2024generating, zhang2023insight}.

A final challenge concerns evaluation. Unlike forecasting accuracy, explanation quality lacks a clear ground truth. Human evaluation by domain experts is expensive and does not scale, while generic text similarity metrics correlate poorly with human judgment \cite{gehrmann2021gem, khapra2021tutorial, dhingra-etal-2019-handling}. As a result, it remains difficult to systematically assess whether generated explanations are genuinely useful.

\paragraph{Problem scope}
\label{sec:overview}
We study the problem of generating \emph{grounded} natural language explanations for time series forecasts. We propose a modular, domain-agnostic framework that takes as input a target time series (TS) with historical observations (TS$_\text{hist}$), its associated forecast (TS$_\text{fc}$), and auxiliary contextual information, such as historical analyst-written explanations (AnX) or external signals (EXT). The objective is to generate a natural language explanation (GenX) that justifies the forecast by linking predicted trends to relevant, verifiable evidence. The problem of generating the forecasts themselves is outside the scope of this study.

\paragraph{Case studies}
To evaluate the proposed framework across distinct settings, we apply it to two domains:

The primary case study focuses on forecasts for the NASDAQ-100 index. This setting provides rich, well-established explanatory narratives and high-quality expert commentary, making it a natural benchmark for grounded explanation generation. 

\begin{figure}[t]
    \centering
    \includegraphics[width=0.95\linewidth]{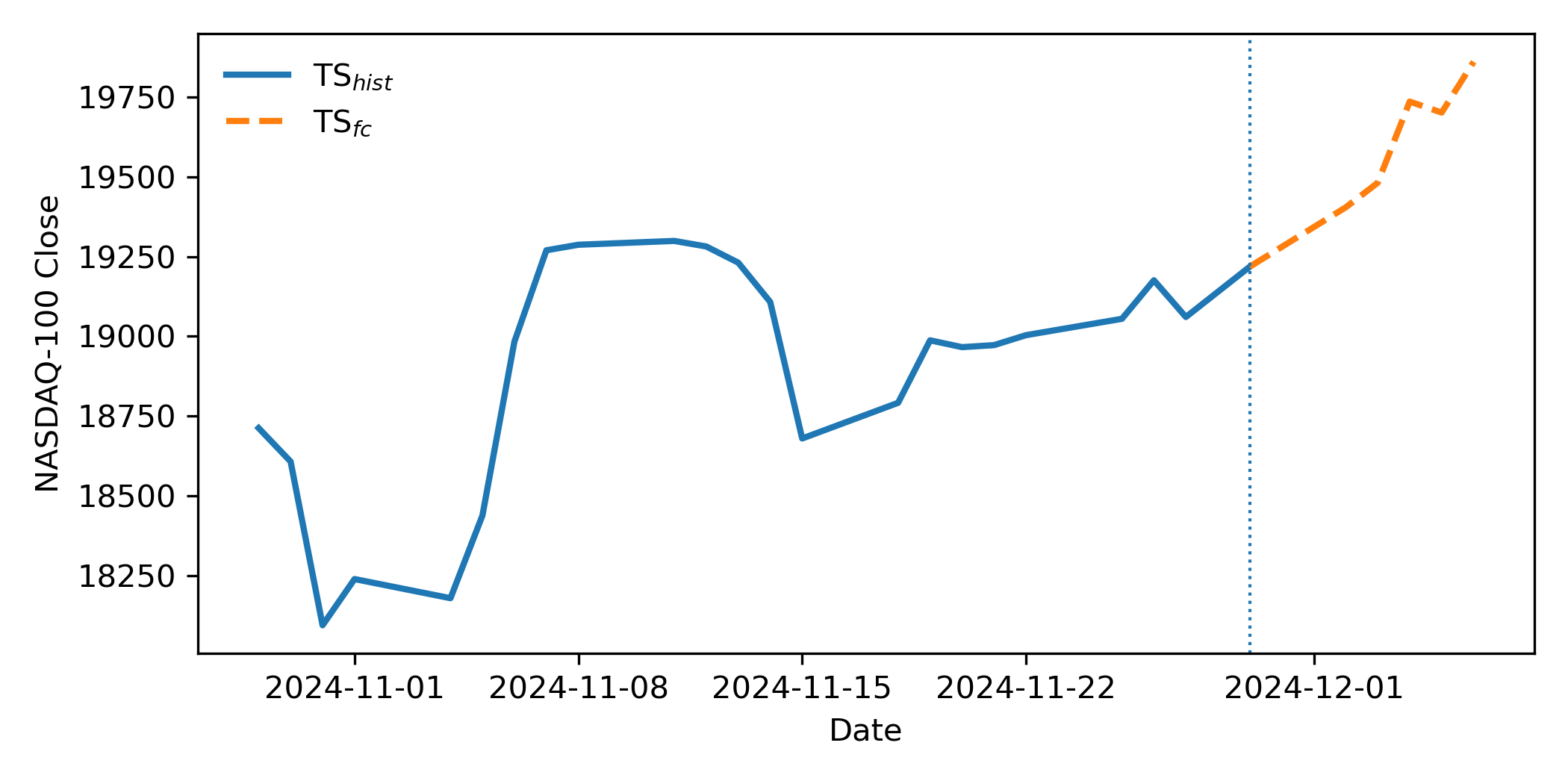}
    \caption{Target time series for the running NASDAQ-100 example. The solid line shows the one-month historical window (TS$_\text{hist}$) ending on the report date (November~29,~2024). The dashed line shows the one-week forecast horizon (TS$_\text{fc}$).}
    \label{fig:nasdaq-ts}
    \Description{Line chart showing the NASDAQ-100 closing price over time. The horizontal axis shows dates from late October to early December 2024, and the vertical axis shows index closing values. A solid line labelled TS hist represents historical daily closing prices over a one-month window. A vertical dotted line marks the report date, separating historical data from the forecast period. To the right of this boundary, a dashed line labelled TS fc shows forecasted closing prices over a one-week horizon.}
\end{figure}

\begin{runningexample}[NASDAQ-100]\label{ex:running}
In this setting, the target time series TS consists of the daily closing price of the NASDAQ-100 index. For a report generated on November~29,~2024, the historical segment TS$_\text{hist}$ spans the preceding one-month window, from October~29 to November~29,~2024. As illustrated in Figure~\ref{fig:nasdaq-ts}, the index exhibits a modest upward trend with some fluctuations over this period. The forecast segment TS$_\text{fc}$ covers the subsequent one-week horizon, corresponding to the five trading days following the report date. 

Auxiliary context is drawn from a corpus of analyst-written market reports (AnX) explaining prior NASDAQ-100 movements, together with aligned external signals (EXT) such as U.S.\ unemployment rates or geopolitical developments during the same period. An appropriate generated explanation (GenX) should justify the predicted increase in TS$_\text{fc}$ by grounding claims in both the observed dynamics of TS$_\text{hist}$ and relevant EXT, rather than merely restating the numerical forecast.
\end{runningexample}

As a complementary case study, we apply the same pipeline to freight-rate forecasting in the shipping industry, using proprietary industry data and expert feedback from Vortexa analysts. Studying these two settings allows us to assess the framework across domains with different data characteristics and decision-making contexts. For clarity and depth of analysis, the remainder of this paper focuses primarily on the NASDAQ-100 case study, with the freight-rate setting used as secondary validation.

\section{Related Work}
\label{sec:related}

Our work lies at the intersection of time series forecasting, explainable machine learning, and LLM–based text generation.

\paragraph{Explainability in Time Series Forecasting}

Explainability has long been recognized as a critical requirement in time series forecasting, particularly in high-stakes domains such as finance and economics \cite{sezer2020financial, bechtel2010forecasting}. Classical statistical models including ARIMA offer limited interpretability through explicit components such as trend and seasonality, but their expressive power is constrained by assumptions of linearity and stationarity \cite{hyndman2018forecasting}.

Modern machine learning models substantially improve predictive performance but often do so at the cost of interpretability. Post hoc explanation techniques, such as feature attribution, saliency methods or attention visualization provide insight into model behaviour but rarely yield explanations that are meaningful to non-technical stakeholders \cite{makridakis2018statistical, jiang2025explainable, triebe2021neuralprophetexplainableforecastingscale} 

In practice, explanations for forecasts are typically provided by human experts, who synthesize numerical patterns with external context such as macroeconomic indicators, events, or policy changes \cite{park2025analysts}.

\paragraph{LLMs for Time Series}

Recent work has explored applying LLMs to time series tasks by fine-tuning pretrained models for forecasting, classification, or anomaly detection \cite{chang2023llm4ts, zhou2023one}. These approaches typically transform numerical time series into token-like representations or attempt to align time series embeddings with language embeddings. While promising in some settings, such methods primarily focus on predictive performance rather than explanation generation. However, LLMs have been widely used for explanation generation in domains such as summarization and scientific reporting \cite{pang2024generating}. Retrieval augmented generation (RAG) is one widely used technique which incorporates documents or passages from a knowledge base into the model input to reduce hallucination \cite{lewis2020retrieval}. While effective for many natural language tasks, RAG remains under-explored in time series explanation. Time series data is continuous, temporally structured, and often high-frequency; naively retrieving raw series or large collections of documents does not by itself ensure coherent, faithful, or temporally consistent explanations \cite{yang2025timerag, ning2025ts}. Addressing this limitation requires principled mechanisms for selecting, encoding, and conditioning on time series evidence.

\paragraph{Evaluation of Natural Language Explanations}

Evaluating natural language explanations is challenging due to their subjective nature and the absence of a single ground truth \cite{gehrmann2021gem}. Human evaluation by domain experts remains the gold standard but is costly, time-consuming, and difficult to scale \cite{khapra2021tutorial}.

Automatic metrics such as BLEU and ROUGE are widely used but correlate poorly with human judgment for explanation tasks \cite{dhingra-etal-2019-handling}. More recent approaches incorporate context-awareness or logical consistency using entailment-based metrics and contradiction detection \cite{bannur2024maira, akyurek2024deductive}. LLMs themselves have also been used as judges to assess explanation quality along qualitative dimensions.

Despite these advances, there is no consensus on how to evaluate explanations in time series settings, and existing studies often rely on ad hoc metrics or small-scale human evaluations.

\section{Architecture}
\label{sec:architecture}

We propose a modular, domain-agnostic framework for generating grounded natural language explanations for time series forecasts. The framework consists of three stages, illustrated in Figure~\ref{fig:teaser}:
\begin{enumerate}
    \item First, we introduce a method for \emph{historical factor extraction}, in which explanatory factors are automatically extracted from AnX and represented in a structured form.

    \item Second, we condition an LLM for \emph{grounded explanation generation} using RAG, in which the model is explicitly conditioned on selected evidence.

    \item Finally, to assess explanation quality at scale, we introduce a \emph{multi-axis evaluation framework} that measures readability, logical consistency, and persuasiveness.
\end{enumerate}

This modular design allows each stage to be analysed independently and adapted across domains, while supporting iterative refinement based on evaluation feedback.

\section{Historical Factor Extraction}
\label{sec:factor_extraction}

\begin{figure}[ht]
    \centering
    \includegraphics[width=0.95\linewidth]{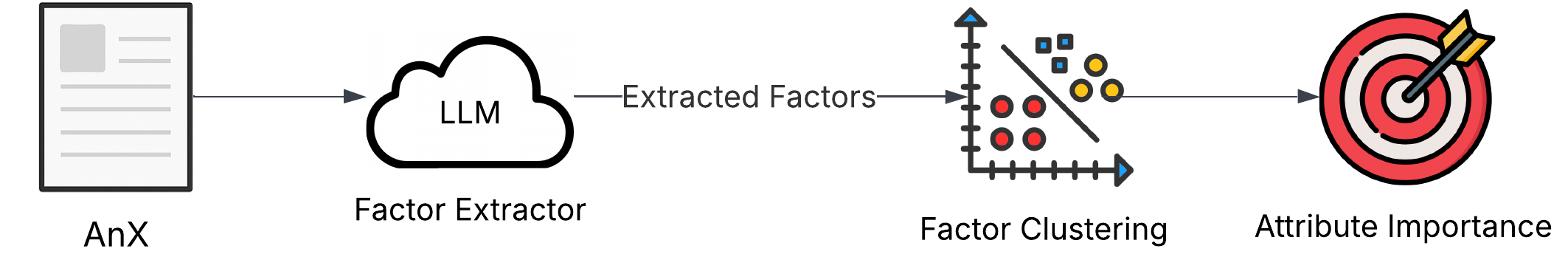}
    \caption{The historical factor extraction pipeline.}
    \label{fig:feature_extraction}
    \Description{Diagram illustrating the historical factor extraction pipeline. On the left, AnX are being fed to an LLM labelled Factor Extractor which outputs extracted factors. The extracted factors are then processed by a Factor Clustering step. The result of the Factor Clustering step is being fed into the final step, labelled Attribute Importance.}
\end{figure}

This section describes the first stage of our framework: \emph{historical factor extraction}. The goal of this stage is to identify and structure the explanatory content that experts use when justifying time series forecasts. We present a domain-agnostic method that learns explanatory structure directly from AnX, without assuming access to labelled explanations or predefined causal variables. An overview of the factor extraction pipeline is shown in Figure~\ref{fig:feature_extraction}.

\subsection{Motivation and Problem Formulation}

Analyst-written explanations are rich sources of information as they rarely describe forecasts in purely numerical terms; instead, they contextualize predicted trends by synthesizing historical patterns, external signals, and anticipated events. Despite stylistic variation, such reports exhibit recurring explanatory elements, for example, references to macroeconomic indicators, sector-level performance, or upcoming earnings announcements.

However, this explanatory structure is not explicit. Forecast explanations are unstructured text, and there is no standardized annotation of which elements constitute causal evidence or how they should be represented. This poses two challenges. First, it is unclear how to reliably extract explanatory content without supervision. Second, even if extracted, it is unclear which aspects of this content are actually informative for grounding explanations.

We frame historical factor extraction as the problem of learning a structured representation of explanatory influences from unstructured text under weak supervision. Given a corpus of historical analyst-written explanations, our goal is to identify a set of factors that (i) are grounded in the source text, (ii) recur across AnX, and (iii) can be used to condition downstream explanation generation.

\paragraph{Factor definition} We define a \emph{factor} as a semantically meaningful cause or influence that is explicitly linked to a forecast in a analyst-written explanation. A factor may reference a trend, an event, or a broader contextual force, provided it plays an explanatory role in shaping the predicted trajectory.

\begin{definition}[Factor]
\label{def:factor}
Formally, a factor is a tuple $f = (n, e, A)$, where $n$ is a concise descriptive name, $e$ is a span of text from the source AnX providing explicit evidence for the factor, and $A$ is a set of attributes that characterize how the factor is used to explain a forecast, including its temporal scope and contextual properties.
\end{definition}

In practice, extracted factors tend to fall into two broad classes. \emph{Quantitative factors} correspond to numerical signals that can be represented as auxiliary time series, such as unemployment rates, inflation indices, or GDP growth. \emph{Contextual factors} correspond to influences that are primarily textual in nature, such as geopolitical events, policy announcements, or firm-specific news. Additionally, we distinguish between \emph{domain-agnostic} and \emph{domain-specific} attributes within a factor's attributes, $A$. Domain-agnostic attributes capture explanatory structure that generalizes across settings - such as the time horizon in which a factor is expected to affect TS. Domain-specific attributes encode application-dependent details (e.g. market-specific conventions for a finance case study) and are treated as optional extensions that enrich explanations within a particular domain without affecting the core framework.

\subsection{Factor Extraction from AnX}
\label{sec:factor-eg}
Extracting factors from free-form text is challenging due to ambiguity, stylistic variation, and the absence of labelled supervision. We address this by leveraging the pattern recognition capabilities of LLMs while constraining their outputs through explicit schemas.

\paragraph{Preprocessing and Chunking.}
AnX are cleaned and segmented into paragraph-aware chunks to ensure that inputs remain within the LLM’s context window while preserving semantic coherence.

\paragraph{LLM-based Extraction.}
Each chunk is passed to an LLM along with a structured prompt specifying the factor definition, required output schema, constraints on evidence attribution, and a small number of illustrative examples. The full prompt is provided in Appendix~\ref{app:prompts}. Crucially, the model is required to associate each factor with an explicit evidence span from the source text. Factors lacking textual support are discarded during post-processing.

\paragraph{Post-processing and Normalization.}
Extracted factors are validated and normalized. Incomplete or malformed records are removed, categorical attributes are standardized using controlled vocabularies, and free-text fields are cleaned to ensure consistency. The output of this stage is a collection of structured factor instances extracted from the historical corpus. 

\begin{runningexample}\label{eg:factor-eg}Consider the following analyst statement extracted from a NASDAQ-100 AnX:
\begin{quote}
    Interestingly, at the macroeconomic level, more U.S. workers are job hunting this July compared to last year, as satisfaction with wages and non-wage benefits at their current jobs has dropped, according to a Federal Reserve Bank of New York survey. The share of people who have searched for a job in the past four weeks jumped to 28.4\%, the highest since March 2014, up from 19.4\% in July 2023.
\end{quote}
From this sentence, we extract a factor capturing increased job hunting as a medium-term macroeconomic influence with a negative expected impact on the index, as shown in Figure~\ref{fig:factor-example}.
\end{runningexample}

\begin{figure}[ht]
\centering
\begin{lstlisting}[language=json,numbers=none]
{
"factor_name": "Increased Job Hunting",
"evidence": "The share of people who have searched for a job in the past four weeks jumped to 28.4%, the highest since March 2014, up from 19.4% in July 2023.",
"attributes" : {
    //domain-agnostic
    "time_horizon": "Medium-term", 
    "time_horizon_value": "next four months",
    "sector_affected": "General", 

    //domain-specific
    "index_component": "Index-wide", 
    "factor_type": "Macroeconomic",
    "price_trend_qualitative": "Bearish",
    "price_trend_quantitative": "Unknown",
    "price_impact_qualitative": "Negative"
    }
}
\end{lstlisting}
\caption{Example of an extracted explanatory factor.}
\label{fig:factor-example}
\end{figure}

\subsection{Factor Consolidation via Clustering}
Different AnX may reference the same underlying explanatory influence using varied phrasing. To identify reusable explanatory patterns in a domain-independent manner, we cluster extracted factors based on semantic similarity.

\paragraph{Embedding and Dimensionality Reduction.}
Factor names are embedded using pretrained sentence embeddings \cite{minilmv2medium2022}. Because clustering in high-dimensional spaces is unreliable due to distance concentration effects \cite{domingos2012few, aggarwal2001surprising}, we apply dimensionality reduction using UMAP \cite{mcinnes2018umap} to preserve local semantic structure.

\paragraph{Density-based Clustering.}
We apply HDBSCAN \cite{mcinnes2017hdbscan} to group semantically similar factors while identifying outliers. The algorithm does not require specifying the number of clusters in advance and naturally handles variable cluster density.

\paragraph{Cluster Labelling.}
Each cluster is labelled using the factor closest to the cluster centroid in embedding space, yielding a set of high-level explanatory themes. The cluster label is then added to the factor as a domain-agnostic attribute.

\begin{runningexample}
The factor extracted in Figure~\ref{fig:factor-example} is assigned to a cluster capturing labour-market dynamics, labelled \emph{``Unemployment Rate and Jobless Claims''}. Similarly, factors such as \emph{``CPI surge''} and \emph{``cost-of-living increases''} are grouped into a separate cluster titled \emph{``Consumer Inflation Expectations''}.
\end{runningexample}
\subsection{Factor Extraction Quality}
Having extracted and clustered factors from AnX, the final stage of the extraction pipeline validates the quality of these outputs and analyses the importance of individual factor attributes. Recall from Definition~\ref{def:factor} that a factor is composed of multiple attributes, not all of which contribute equally to explanatory reasoning. While clustering groups factors into semantically salient themes, downstream explanation generation requires selecting a compact subset of high-signal attributes. This is necessary to avoid conditioning LLMs on redundant context, reducing context overload.


 \begin{figure}[!ht]
    \centering
    \includegraphics[width=0.99\linewidth]{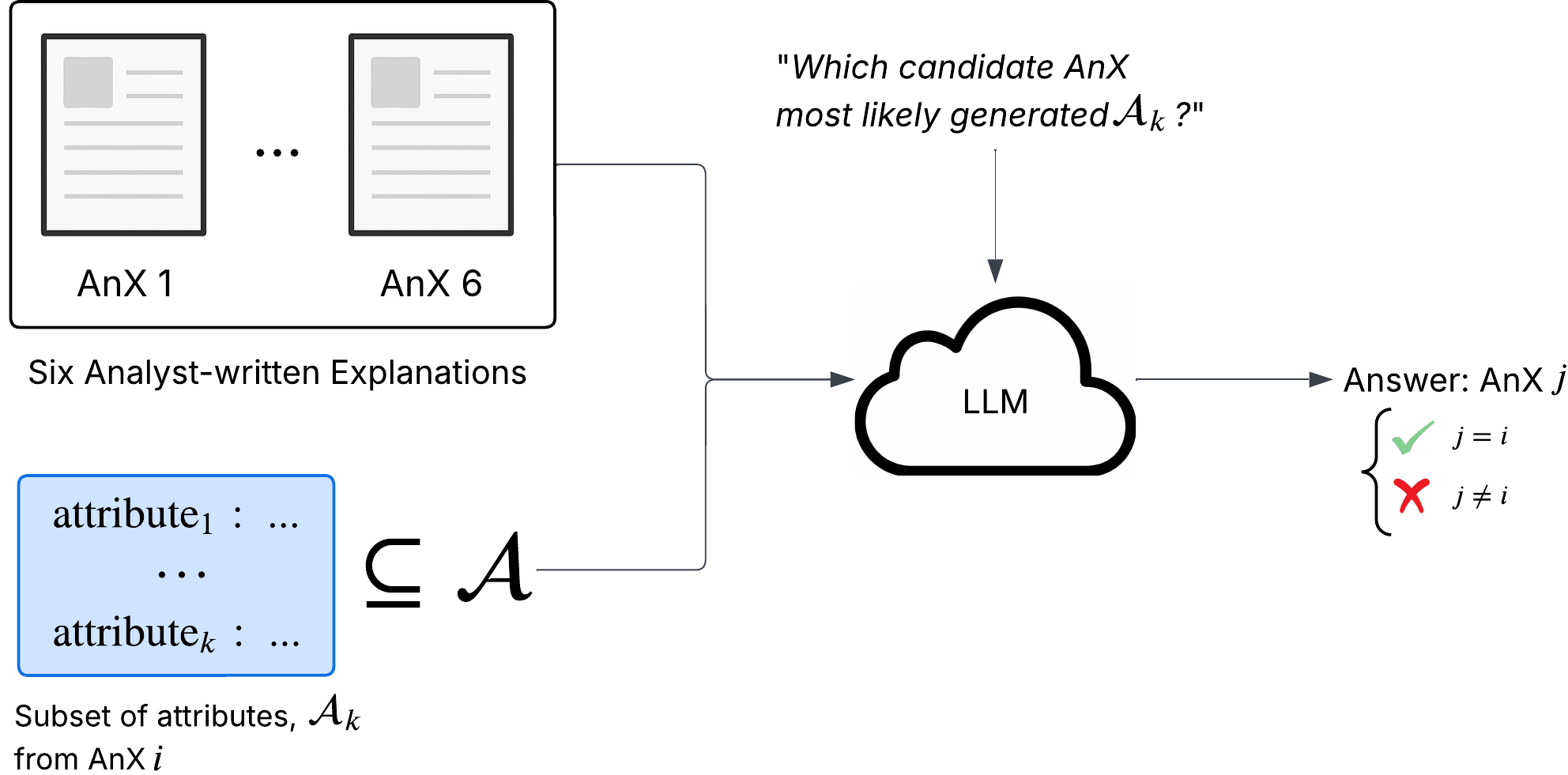}
    \caption{The AnX retrieval pipeline. A reasoning model is asked to identify the source AnX from which a subset of factor attributes were extracted.}
    \label{fig:etiological_reasoning_setup}
    \Description{On the left, there is a set of six analyst-written explanations labelled AnX 1 through AnX 6. Below them, a highlighted box represents a subset of attributes A_k extracted from one explanation AnX i. Both the set of candidate explanations and the attribute subset are provided as input to an LLM. Above the LLM, a question asks which candidate explanation most likely generated A_k. On the right, the LLM outputs a selected explanation AnX j, with a visual indication of a correct match when j equals i and an incorrect match when j does not equal i.}

\end{figure}

\paragraph{AnX Retrieval Task.}
We first introduce the retrieval task, a diagnostic subproblem inspired by etiological reasoning, and then describe its use in validating extraction quality and identifying which attributes should be retained for downstream conditioning (Figure~\ref{fig:etiological_reasoning_setup}). The task poses the following question to a reasoning LLM: \textit{Given a set of analyst-written explanations, $\{\text{AnX}_1, \ldots, \text{AnX}_m\}$, and a partial set of attributes $A_k$ describing factors from one of these AnX, can the model correctly infer which AnX those attributes were extracted from?}

Formally, we sample a set of analyst-written explanations $\mathcal{R} = \{\text{AnX}_1, \ldots, \text{AnX}_m\}$ and pick an $i \in [1,\ldots,m]$. Let the factors extracted from $\text{AnX}_i$ be $F_i$. From each $(n^j, e^j, A^j)_{j \in [1, \ldots, |F_i|]} \in F_i$, we select a subset of attributes $\mathcal{A}^j_k$ (e.g. \{\texttt{factor\_cluster}, \texttt{time\_horizon}\}) and frame the task as an $m$-way classification problem: 

Given $\{\mathcal{A}^1_k, \ldots,\mathcal{A}^{|F_i|}_k\}$ and $\{\text{AnX}_1, \ldots, \text{AnX}_m\}$, the model must select the correct source $\text{AnX}_i$ from a set of $m$ candidates. For our experiments, we set $m=6$.

\begin{runningexample}
Let the analyst-written explanation excerpted in Section~\ref{sec:factor-eg} be AnX$_1$, and take five additional randomly sampled analyst-written explanations, $[\text{AnX}_2, \ldots, \text{AnX}_6]$ from the corpus. Suppose that from the factor extracted from $\text{AnX}_1$, we retain only a low-signal attribute such as \texttt{\{time\_horizon\_value : ``next four months''\}}. Given this information alone, a reasoning model is unlikely to reliably identify the correct source AnX, as similar temporal descriptors recur across many explanations.

In contrast, if the model is instead provided with a higher-signal attribute such as \texttt{\{factor\_cluster: ``Unemployment Rate and Jobless Claims''\}}, the task becomes substantially easier. Such attributes capture more AnX specific content, enabling the model to more accurately infer which AnX the factors were extracted from.
\end{runningexample}

\paragraph{Validation via Retrieval.}
We first apply the retrieval task using the \emph{full set of extracted factor attributes} ($A_k = A$). Retrieval accuracy in this setting serves as a direct measure of extraction quality: if the factors encode meaningful explanatory content, a reasoning model should be able to reliably map them back to their source explanations. Conversely, if extraction yields generic or noisy attributes, performance should degrade toward chance.

\paragraph{Attribute Importance via Forward Variable Selection.}
While retrieval with all attributes validates extraction quality in aggregate, it does not indicate which attributes are responsible for the observed performance. To quantify attribute importance, we reuse the retrieval task as a supervised proxy objective and apply a forward variable selection procedure \cite{guyon2003introduction}, analogous to feature selection in predictive modelling. Starting from single-attribute subsets, we iteratively add the attribute that yields the largest improvement in retrieval accuracy to a working set $\mathcal{T}$. The process continues until no additional attribute produces a statistically significant gain. The resulting set $\mathcal{T}$ represents a compact combination of attributes that captures most of the explanatory signal and is used to define the subset of attributes used for downstream explanation generation.




\section{Grounded Explanation Generation}
\label{sec:grounded_generation}
The second stage of our framework focuses on generating natural language explanations (GenX) for time series forecasts that are both coherent and explicitly grounded in data-derived evidence. Building on the structured factors identified in Section~\ref{sec:factor_extraction}, the goal of this stage is not only to produce fluent GenX, but to ensure that they satisfy four key requirements: (i) narrative coherence, (ii) grounding in observable evidence, (iii) temporal consistency, and (iv) relevance and brevity.

\begin{figure}[!ht]
    \centering
    \includegraphics[width=0.99\linewidth]{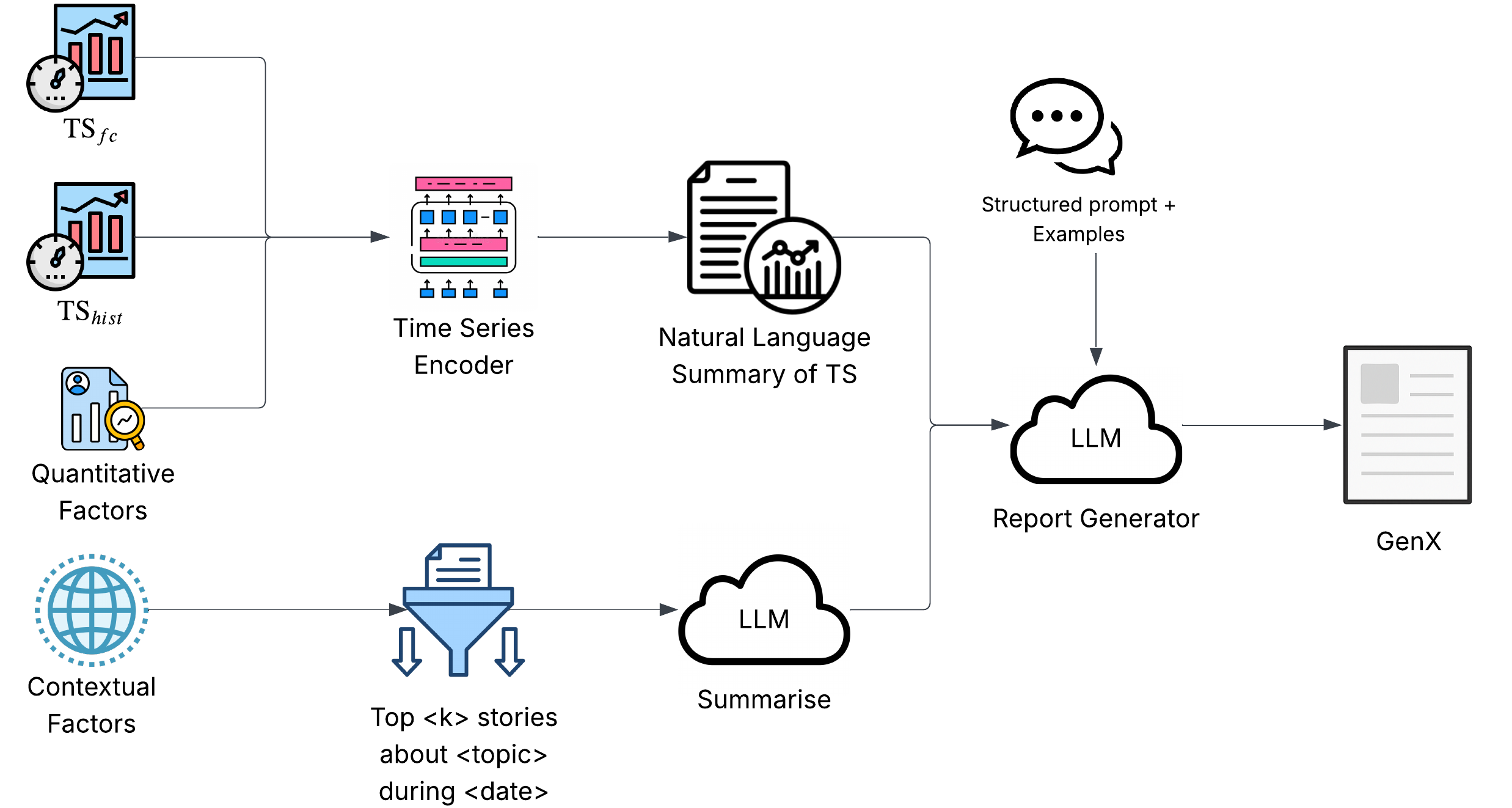}
    \caption{The grounded explanation generation stage.}
    \label{fig:report_gen_overview}
    \Description{On the left, TS hist and TS fc and quantitative factors are provided as input to a time series encoder, which produces a natural language summary of the TS. In parallel, contextual factors are fed into a filter that retrieves 'Top <k> stories about <topic> during <date>,. This is fed into a summariser LLM. All inputs are then combined and inputed, together with a structured prompt and example reports to an LL report generator. On the right, the report generator outputs a generated report.}

\end{figure}

Figure~\ref{fig:report_gen_overview} provides an overview of the grounded explanation generation stage. The pipeline consists of three main components: evidence retrieval, evidence encoding, and conditional generation.

\paragraph{Evidence Retrieval}
Evidence retrieval operationalizes the factors identified in Section~\ref{sec:factor_extraction} by mapping each identified factor cluster to concrete evidence in EXT. For each cluster, we associate either quantitative evidence (e.g. aligned auxiliary time series such as unemployment rates or inflation indices) or qualitative evidence (e.g. news articles or event summaries) that instantiate the underlying explanatory theme. This mapping is domain-specific and partially human-assisted as it relies on domain knowledge and conventions that are difficult to fully automate.

\begin{runningexample}
Consider the factor cluster \emph{``Unemployment Rate and Jobless Claims''}. In the evidence retrieval step, this cluster is mapped to quantitative labour-market signals, including a weekly U.S.\ unemployment rate time series and initial jobless claims data. These time series are retrieved over a broad temporal range sufficient to support report generation, and the relevant subsections are selected based on the report date and the historical target time series window.
\end{runningexample}

\paragraph{Encoding Time Series}
Numerical time series data cannot be directly consumed by LLMs in a form that supports reliable reasoning. We therefore encode time series into compact, interpretable textual representations that preserve salient temporal structure while avoiding raw numeric overload. This encoding procedure is applied not only to $\text{TS}_{hist}$ and $\text{TS}_{fc}$, but also to quantitative EXT (e.g. unemployment rates).

\begin{enumerate}
\item{\emph{Multi-scale Representation.}}
To capture both short-term dynamics and long-term trends, time series are represented at multiple resolutions (e.g. daily, weekly, and monthly).

\item{\emph{STL Decomposition.}}
At each resolution, we apply seasonal-trend decomposition to separate trend, seasonal, and residual components \cite{cleveland1990stl}. Additionally, we compute the magnitude and direction of changes at each resolution (e.g. day-on-day, week-on-week, or month-on-month).

\item{\emph{Natural Language Summarization.}}
The resulting statistics are converted into concise text dictionaries (e.g. describing the direction and strength of trends or presence of recurring cycles), enabling the LLM to integrate temporal signals into broader causal reasoning while maintaining traceability to the underlying data \cite{tang2025llmpsempoweringlargelanguage}.
\end{enumerate}

\begin{runningexample}
In our running example, unemployment rate is encoded at daily, weekly, and monthly resolutions. After STL decomposition, the resulting summaries capture a steadily increasing short-term trend, limited seasonal structure, and a positive week-over-week change magnitude. The final dictionary provided to the LLM is illustrated in Figure~\ref{fig:ts-encoding}.
\end{runningexample}

\begin{figure}[!ht]
\begin{lstlisting}[language=json,numbers=none]
{
"metric": "Unemployment Rate"
"units": "% of labour force"
"sampling_frequency": "weekly"
"trend_gradient_from_2024-07-07_to_2024-12-01": +0.664, 
"trend_gradient_description_from_2024-07-07_to_2024-12-01": "steady positive", 
"latest_percentage_week_on_week_change": +0.09, 
"latest_absolute_week_on_week_change": +6.832,
"weekly_seasonality": "None"
}
\end{lstlisting}
    \caption{Example encoding for weekly Unemployment Rate}
    \label{fig:ts-encoding}
\end{figure}

\paragraph{Integrating External Context}
In many forecasting settings, explanations rely not only on internal time series signals but also on external contextual information, such as policy changes or geopolitical instability. To support such reasoning, we incorporate external context into the conditioning pipeline. Contextual factors extracted from AnX are used as queries to retrieve related news articles and documents. An LLM is prompted to summarize retrieved text into compact, fact-preserving statements that emphasize entities, temporal markers, and salient developments.

\begin{runningexample}
In our running example, one contextual factor cluster is ``Middle East Tensions''. We use this to retrieve the top 10 news articles about tensions in the Middle East up until the dat of report generation. These articles are summarized into short textual statements highlighting key facts (e.g.recent policy changes), which are then provided to the explanation generator.
\end{runningexample}

\paragraph{Structured Prompting}
All encoded evidence is interleaved into a structured prompt that conditions the LLM during generation, as shown in Figure~\ref{fig:report_gen_overview}. The full prompt is given in Appendix~\ref{app:prompts}.
We use in-context learning to control output structure and style, embedding a small number of exemplars that demonstrate desired narrative conventions \cite{dong2022survey}. Importantly, these examples serve to anchor tone and structure rather than to provide substantive content, reducing the risk of template copying or over-fitting.
The structured prompting approach constrains the model’s degrees of freedom, reducing hallucination and improving logical consistency. The LLM acts primarily as a linguistic realizer, aggregating and articulating evidence supplied by upstream components rather than inferring explanations from latent knowledge alone.\\

Empirically, this structured conditioning approach plays a central role in explanation quality. Encoding time series into semantically meaningful summaries improves temporal grounding and substantially reduces internal contradictions, while selectively integrated external news enables explanations to link numerical trends to plausible real-world drivers. Together, these components constrain generation to high-signal evidence, improving logical consistency and persuasiveness without sacrificing fluency. A complete example of a generated explanation is provided in Appendix~\ref{app:generated_report}.



\section{Evaluation and Refinement}
\label{sec:evaluation}

This section presents a scalable evaluation framework for assessing the quality of GenX without relying on extensive expert judgment. Existing evaluation approaches for natural language generation are limited in three respects: (i) many automated metrics are generic and do not account for domain specialization; (ii) metrics often measure surface-level fluency rather than deeper properties such as internal consistency and argumentative strength; and (iii) when deeper properties are evaluated, expert human assessment is typically required, which is costly and difficult to scale. We address these limitations by evaluating explanations along three complementary axes: readability, consistency in reasoning, and persuasiveness. In addition to benchmarking, these signals can be used for iterative refinement of the generation pipeline.

\paragraph{Readability}

Readability measures whether an explanation communicates its analysis clearly and fluently while maintaining an appropriate professional register. We evaluate readability using both a generic index and a domain-sensitive measure.

\begin{enumerate}
\item{\emph{General readability.}}
We use the Gunning--Fog Index (GFI) as a coarse measure of general readability \cite{gunning1952technique}.
\[
\text{GFI} = 0.4 \times \left( \frac{\text{total words}}{\text{total sentences}} + 100 \times \frac{\text{complex words}}{\text{total words}} \right).
\]
GFI uses syllable-based complexity, which may not always reflect actual comprehensibility, particularly for domain-specific terminology. Thus, we treat GFI as only a rough indicator of stylistic complexity.

\item{\emph{Domain-specific readability.}}
To measure domain-specific readability, we employ cloze tests \cite{taylor1953cloze}, where key terms are removed and a reader must fill in the blanks. Rather than using human readers, we use an LLM to answer each cloze question, and report accuracy as a proxy for how predictable and contextually appropriate the specialized terminology is.

The effectiveness of cloze tests depends on selecting informative blanks and plausible distractors. We adopt a \emph{smart cloze} procedure adapted from \cite{redmiles2019comparing}. Text is part-of-speech tagged, and only content-bearing tokens (e.g. nouns, adjectives/adverbs, directional descriptors) are eligible to be blanked. Distractors are sampled from a part-of-speech--matched vocabulary so that all options are grammatically plausible but semantically mismatched. This yields a domain-sensitive readability signal that rewards coherent and context-appropriate usage of specialized terms.
\end{enumerate}

\paragraph{Consistency in Reasoning}
Beyond readability, explanations must be internally consistent: contradictions undermine trust and can degrade downstream decision-making \cite{long2023contradictions}. We evaluate consistency using a contradiction-detection procedure inspired by deductive closure training \cite{akyurek2024deductive}.

First, each explanation is converted into a set of assertions with a canonical form and vocabulary (e.g. \textit{``\{change\} in \{subject\} caused by \{reason\}''}) using an LLM.

We then construct a small set of seed documents consisting of domain axioms and general purpose axioms, also encoded in canonical form. Each extracted assertion is evaluated independently against these axioms by an LLM acting as a proxy reasoner. The model outputs a binary label indicating if the assertion can be true given the axioms, together with log-probabilities indicating its confidence in the label. High-confidence \textit{False} labels are treated as contradictions, while low-confidence cases are flagged as ambiguous.

\begin{runningexample}
    As an example of a domain-specific axiom in the financial setting, we include constraints encoding well-established macroeconomic relationships. For instance, an axiom may state that, all else being equal, a sustained increase in interest rates exerts downward pressure on index valuations. In canonical form, this can be represented as \textit{``decrease in index prices can be caused by sustained increase in interest rates''} and \textit{``increase in index prices cannot be caused by sustained increase in interest rates''}. An extracted assertion claiming that rising interest rates directly caused a rise in index prices would therefore be flagged as contradictory under the second axiom.

\end{runningexample}
\paragraph{Persuasiveness}
Persuasiveness measures whether an explanation presents evidence and claims in a convincing way. Persuasiveness is inherently subjective and context-dependent, and there is no universally accepted scalable metric that reliably captures human judgment \cite{borah2025persuasion, rogiers2024persuasion}. We therefore use an LLM-as-a-judge setup to approximate relative persuasiveness in a scalable manner \cite{gu2024survey}.

Given a pair of explanations describing the same context (e.g. a GenX and a reference AnX for the same TS), the judge model decides which is more persuasive, or if both are equally persuasive. To mitigate positional bias, both orderings of each pair are evaluated and aggregated.

Because LLM judges may be sensitive to superficial cues (e.g. length or style), we validate the judge using controlled paraphrasing: a separate model paraphrases analyst-written explanations while preserving factual content and structure, and the judge should label most original--paraphrase pairs as ties. This sanity check provides evidence that the judge responds primarily to substantive differences rather than surface variation.

\paragraph{Using Evaluation Signals for Refinement}
\label{sec:refinement}
Evaluation is not only used for benchmarking but can provide actionable feedback for iterative refinement, as illustrated by the feedback loop in Figure~\ref{fig:teaser}. In particular, contradiction detection is incorporated into a regeneration loop: propositions flagged as contradictory with high confidence are returned to the generator as explicit constraints, prompting revision to eliminate inconsistencies. More generally, prompt-level interventions are used to reduce contradictions and improve argument structure. Contrastive prompting \cite{chia2023contrastive} exposes models to both consistent and inconsistent reasoning patterns, while self-check instructions encourage models to verify their claims \cite{miao2023selfcheckusingllmszeroshot}.

\section{Case Study: NASDAQ-100 Index}
\label{sec:nasdaq_case_study}
This section provides a detailed evaluation of the proposed framework on our NASDAQ-100 case study. This domain provides a challenging testbed due to the interaction of macroeconomic, sectoral, and firm-level drivers, as well as the availability of high-quality AnX that enables weakly supervised learning.

\paragraph{Experimental Setup}
We apply the full pipeline described in Sections~\ref{sec:factor_extraction}--\ref{sec:evaluation}. To study the effect of model reasoning capacity on explanation quality, we generate explanations using two LLMs: \texttt{gpt-4o-mini}, which serves as a lightweight baseline, and \texttt{o3}, a stronger reasoning model. Both models use the same conditioning schema and prompts, isolating the impact of the underlying model rather than differences in input information. Using the latest 100 AnX retrieved from the NASDAQ portal as of July~23,~2025 as a reference set, we generate 100 explanations for the same weeks with \texttt{gpt-4o-mini}, and a smaller sample of 5 explanations using \texttt{o3} due to resource constraints. GenX are evaluated against AnX using the automated evaluation framework described in Section~\ref{sec:evaluation}.

\subsection{Findings}
\begin{table*}[!t]
\centering
\begin{tabularx}{0.95\linewidth}{X|X|X XXXX}
\toprule
\multicolumn{3}{c}{\small{\textbf{Configuration}}} & \multirow{2}{=}{\small{\textbf{Gunning Fog Index} ($\downarrow$)}} & \multirow{2}{=}{\small{\textbf{Smart Cloze} ($\uparrow$)}} & \multirow{2}{=}{\small{\textbf{Contradictions} ($\downarrow$)}} & \multirow{2}{=}{\small{\textbf{Persuasiveness} ($\uparrow$)}} \\
\cmidrule{1-3}
\small{Time-Series Encoding}&\small{External News}&\small{Examples} \\
\midrule
\xmark & \xmark & \xmark& 17.5 & 45.2 & 0.087 & -0.72\\
\cmark & \xmark & \xmark& 17.5 & 45.3 & \textbf{0.065} & \textbf{-0.54}\\
\xmark & \cmark & \xmark& 17.6 & 47.3 & 0.086 & \underline{-0.60}\\
\xmark & \xmark & \cmark & \textbf{16.5} & \textbf{51.3} & 0.090 & -0.67\\
\cmark & \cmark & \cmark & \underline{16.9} & \underline{50.9} & \underline{0.066} & \textbf{-0.54}\\
\bottomrule
\end{tabularx}
\caption{Impact of Time Series Encoding, External News Summaries and Few-Shot Examples on explanation quality. $\downarrow$ indicates lower scores are better, and $\uparrow$ indicates higher scores are better. Best results for each metric in \textbf{bold}, second-best \underline{underlined}.}
\label{tab:ablation}
\end{table*}
\paragraph{F1: GenX match AnX in readability.}
\begin{figure}[!ht]
    \centering
    \includegraphics[width=0.95\linewidth]{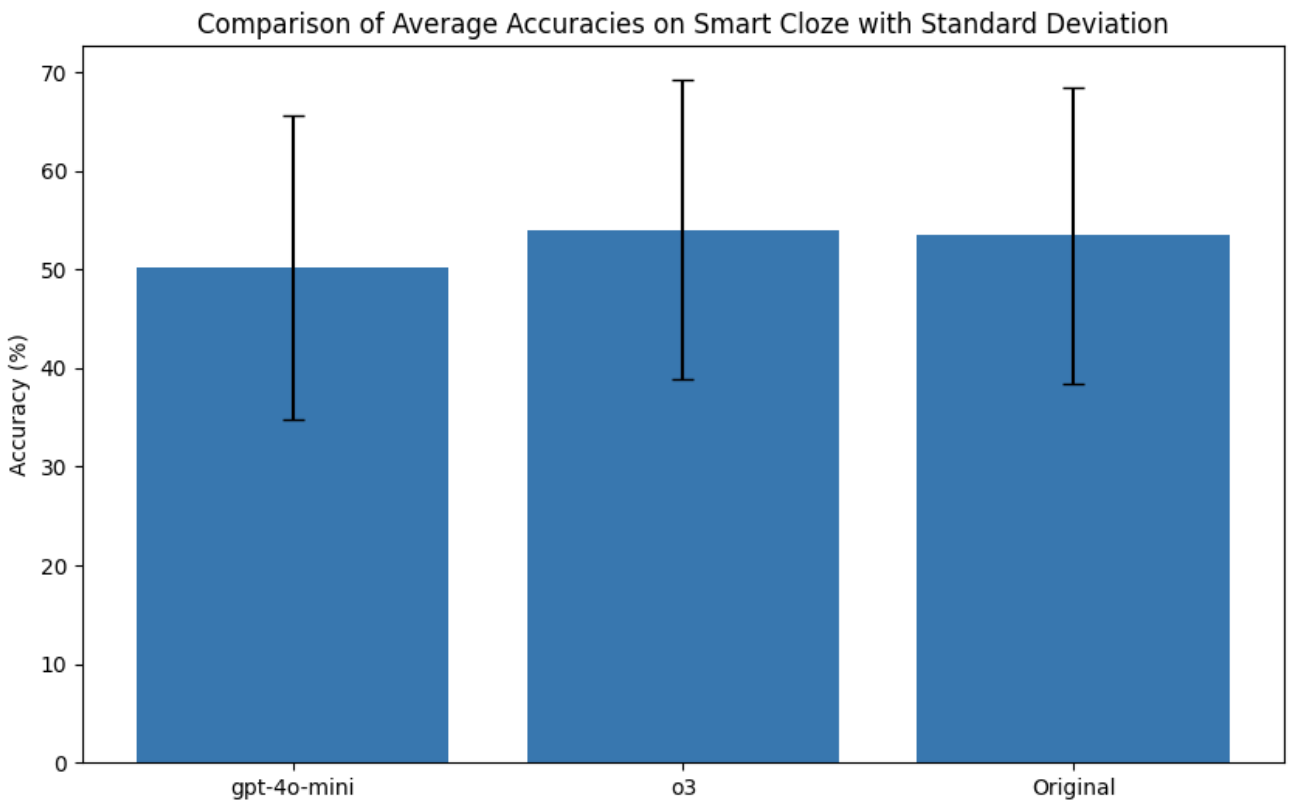}
    \caption{Domain-specific readability comparison using the smart cloze evaluation.}
    \label{fig:smart_cloze_results}
    \Description{Bar chart comparing average accuracy on the smart cloze evaluation across three explanation sources. The horizontal axis lists three categories: gpt-4o-mini, o3, and Original. The vertical axis shows accuracy percentages ranging from 0 to 70 percent. Each category is represented by a single bar indicating mean accuracy, with vertical error bars showing standard deviation.}

\end{figure}
Across both readability metrics, generated explanations are comparable to analyst-written explanations. GFI scores indicate that all explanations fall within an advanced reading range appropriate for professional financial analysis, with no practically meaningful differences between sources. Domain-specific readability, measured via the smart cloze evaluation (Figure~\ref{fig:smart_cloze_results}), shows similarly narrow performance gaps, suggesting that grounded conditioning preserves appropriate use of industry terminology and contextual cues.

\paragraph{F2: Reasoning capacity improves internal consistency.}
\begin{figure}[!ht]
    \centering
    \includegraphics[width=0.95\linewidth]{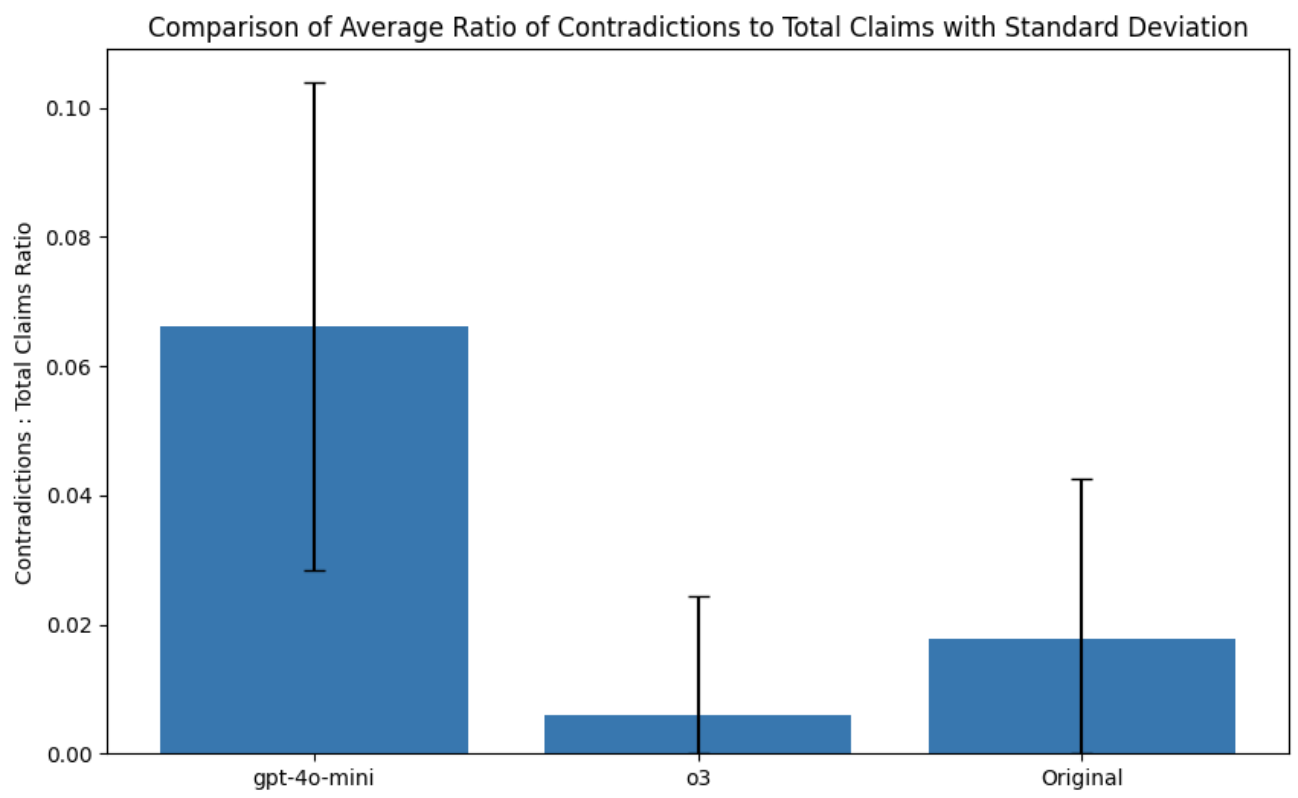}
    \caption{Contradiction rates across explanation types.}
    \label{fig:contradiction_results}
    \Description{Bar chart comparing the average ratio of contradictions to total claims across three explanation sources. The horizontal axis lists gpt-4o-mini, o3, and Original. The vertical axis shows the ratio of contradictions to total claims, ranging from zero to approximately 0.11. Each category is represented by a single bar indicating the mean ratio, with vertical error bars showing the standard deviation.}

\end{figure}
Consistency analysis based on deductive closure reveals clear differences across models (Figure~\ref{fig:contradiction_results}). Explanations generated by \texttt{gpt-4o-mini} exhibit higher rates of internal contradictions than both analyst-written explanations and those generated by the stronger reasoning model \texttt{o3}. While AnX remain the most consistent overall, \texttt{o3} substantially narrows this gap, highlighting the role of reasoning capacity in maintaining logical coherence.

\paragraph{F3: Reasoning capacity improves persuasiveness.}
\newcolumntype{b}{X}
\newcolumntype{s}{>{\hsize=.6\hsize}X}
\begin{table}[ht]
\centering
\begin{tabularx}{0.95\linewidth}{b | sss}
\toprule
\textbf{Comparison} & \textbf{Generated Preferred} & \textbf{Tie} & \textbf{Reference Preferred} \\
\midrule
\texttt{4om} vs Analyst & 15\% & 15\% & 70\% \\
\texttt{o3} vs Analyst & 50\% & 50\% & 0\% \\
\texttt{o3} vs \texttt{4om} & 70\% & 20\% & 10\% \\
\bottomrule
\end{tabularx}
\caption{Persuasiveness evaluation using an LLM. \texttt{4om} means \texttt{gpt-4o-mini}. Percentages indicate the fraction of comparisons where each explanation was preferred or judged equal.}
\label{tab:persuasiveness}
\end{table}

As shown in Table ~\ref{tab:persuasiveness}, persuasiveness evaluation using an LLM-as-a-judge framework shows that analyst-written explanations are generally preferred over \texttt{gpt-4o-mini} outputs, reflecting the depth and rhetorical structure of expert-authored analysis. Explanations generated by the stronger reasoning model \texttt{o3} substantially narrow this gap: in pairwise comparisons against AnX, \texttt{o3} outputs are judged equally persuasive or preferred in majority of cases. We note that this observation is based on a limited sample of \texttt{o3} generated explanations and should therefore be interpreted with caution. 

\paragraph{F4: Ablation analysis reveals complementary effects and context trade-offs.}
The ablation study in Table~\ref{tab:ablation} indicates that no single configuration uniformly dominates across all evaluation dimensions. Each component of the generation pipeline contributes to explanation quality in a distinct way. TS encoding yields the largest reductions in internal contradictions, highlighting its importance for logical and temporal consistency. External news summaries primarily improve persuasiveness by grounding explanations in real-world events, while examples have the strongest effect on stylistic alignment and domain-specific readability.

When all components are enabled simultaneously, the resulting configuration performs strongly across all evaluation dimensions, but does not consistently achieve the best score on every individual metric. Instead, it remains within the top two configurations throughout, reflecting inherent trade-offs between readability, logical consistency, and persuasiveness. One contributing factor may be context saturation. In the full configuration, the model must attend to a larger and more heterogeneous collection of inputs. While this richer context supports grounding and overall coherence, it can diffuse attention across signals.

\paragraph{Human Evaluation}
In addition to automated evaluation, we conducted a small-scale human evaluation to qualitatively assess explanation quality. For the NASDAQ-100 case study, financial domain experts provided structured feedback on GenX, and generally found the explanations to be coherent, well-grounded, and comparable in style to AnX. We further tested generality in a separate freight-rate forecasting setting, where Vortexa analysts performed pairwise comparisons and expressed roughly equal preference between explanations generated by \texttt{o3} and analyst-written explanations. Full details of the human evaluation protocol and results are provided in Appendix~\ref{app:human_eval}.


\section{Conclusion}

We studied the problem of generating grounded natural language explanations for time series forecasts and proposed a modular, domain-agnostic framework that combines structured factor extraction, compact evidence conditioning, and scalable evaluation. By learning explanatory structure from historical expert narratives and using it to constrain LLMs during generation, the framework produces explanations that are comparable to analyst-written explanations across readability, logical consistency, and persuasiveness.

Beyond the financial and freight-rate case studies presented here, the framework provides a general blueprint for producing transparent, evidence-backed explanations in a wide range of decision-support settings, including logistics, supply-chain forecasting, epidemiological surveillance, and infrastructure monitoring.

Several limitations remain. Our evaluation covers only two application domains, and explanation quality depends on the availability and fidelity of structured historical signals. In addition, parts of the evaluation rely on LLM-based judges, which, while scalable, remain imperfect proxies for expert human judgment, and results involving the strongest reasoning model are based on limited samples.

Future work includes reinforcement-based fine-tuning to directly optimize explanation quality, adaptive attribute selection to improve robustness across domains, and extensions to multimodal settings where time series interact with richer external signals.

\bibliographystyle{ACM-Reference-Format}
\bibliography{main}

\appendix 
\section{Prompts}\label{app:prompts}
This appendix lists the full prompts used in this paper.

\subsection{Factor Extraction Prompt}

\begin{lstlisting}[breaklines=true, basicstyle=\ttfamily]
You are an expert in financial markets and equity index analysis. Your task is to analyze the following market report and extract all relevant factors that may affect current and future movements of the **NASDAQ index**.

You should return your answer as text that can be saved as a JSON lines file only, with **no formatting, commentary or explanation**. Ensure each json line is separated by a single new line. Use the following JSON keys in this exact order:

factor_name, evidence, sector_affected, index_component, factor_type, time_horizon, time_horizon_value, price_trend_qualitative, price_trend_quantitative, price_impact_qualitative, price_impact_quantitative

### Factor Classification

**Factor Type**
- `Microeconomic`: related to sector or company specific events, reports, etc.
- `Macroeconomic`: related to macroeconomic events, policy changes, global news, geopolitical risks, interest rates, or inflation data.

**Factor Name**
- Generate a concise and descriptive name for the factor based on your understanding.

**Index Component**
- If a specific NASDAQ-listed company or ticker is involved, the ticker symbol. This should be one of ["AAPL", "ABNB", "ADBE", "ADI", "ADP", "ADSK", "AEP", "AMAT", "AMD", "AMGN", "AMZN", "APP", "ARM", "ASML", "AVGO", "AXON", "AZN", "BIIB", "BKNG", "BKR", "CCEP", "CDNS", "CDW", "CEG", "CHTR", "CMCSA", "COST", "CPRT", "CRWD", "CSCO", "CSGP", "CSX", "CTAS", "CTSH", "DASH", "DDOG", "DXCM", "EA", "EXC", "FANG", "FAST", "FTNT", "GEHC", "GFS", "GILD", "GOOG", "GOOGL", "HON", "IDXX", "INTC", "INTU", "ISRG", "KDP", "KHC", "KLAC", "LIN", "LRCX", "LULU", "MAR", "MCHP", "MDLZ", "MELI", "META", "MNST", "MRVL", "MSFT", "MSTR", "MU", "NFLX", "NVDA", "NXPI", "ODFL", "ON", "ORLY", "PANW", "PAYX", "PCAR", "PDD", "PEP", "PLTR", "PYPL", "QCOM", "REGN", "ROP", "ROST", "SBUX", "SHOP", "SNPS", "TEAM", "TMUS", "TSLA", "TTD", "TTWO", "TXN", "VRSK", "VRTX", "WBD", "WDAY", "XEL", "ZS"]. If not mentioned, use `Index-wide`.

### Market Drivers

**Sector Affected**
- If relevant, specify the sector (e.g. `Technology`, `Healthcare`, `Consumer Discretionary`). If not mentioned, use `General`.

### Temporal Classification

**Time Horizon**
- Choose from:
  - `Short-term`: within 1 week
  - `Medium-term`: 1 week - 1 month
  - `Long-term`: over 1 month
  - `Uncertain`: timing unclear

**Time Horizon Value**
- Include the specific time reference (e.g. `next week`, `Q2 2024`, `after FOMC meeting`). If not mentioned, use `Unknown`.

### Price Information

**Price Trend (Qualitative and Quantitative)**
- `price_trend_qualitative`: Choose from `Bullish`, `Neutral`, or `Bearish`
- `price_trend_quantitative`: Extract any numeric or percentage change mentioned (e.g. `+2.3%`, `-150 points`). If not mentioned, use `Unknown`.
- `price_impact_quantitative`: Provide any numeric estimate of index change or implied volatility (e.g. `+200 pts`, `-1.8%`, `volatility spike`). If not mentioned, use `Unknown`.

### Evidence and Company Information

**Evidence**
- Provide a direct sentence or phrase from the report that supports the factor, wrapped in double-quotation marks.

If you cannot identify any valid factors in the report, return empty text.

Below is an example of a hypothetical report, and JSON formatted factors extracted from the report. You should follow the style of this example.
{example}
\end{lstlisting}

\subsection{Feature Importance Prompt}

\begin{lstlisting}[breaklines=true, basicstyle=\ttfamily]
You are a trader with deep expertise in the NASDAQ-100 index. I have extracted several factors affecting the index from a market report, and I need your help to determine which of the candidate reports these factors most likely came from. Each factor consists of the following features:
    
### Factor Classification

**Factor Name**
- a concise and descriptive name for the factor based on your understanding.

**Index Component**
- ticker of any NASDAQ-listed companies or tickers involved, or `Index-wide`.

### Market Drivers

**Sector Affected**
- the sector (e.g. `Technology`, `Healthcare`, `Consumer Discretionary`) if relevant, or 'general'.

**Microeconomic Driver**
- the specific microeconomic driver (e.g. earnings report, product launch), or blank if factor is macroeconomic.

**Macroeconomic Driver**
- the specific macroeconomic driver (e.g. interest rate change, inflation data), or blank if factor is microeconomic.

### Temporal Classification

**Time Horizon**
  - `Short-term`: within 1 week
  - `Medium-term`: 1 week - 1 month
  - `Long-term`: over 1 month
  - `Uncertain`: timing unclear

**Time Horizon Value**
- Any specific time references, or `Unknown`.

### Price Information

**Price Trend (Qualitative and Quantitative)**
- `price_trend_qualitative`: Bullish`, `Neutral`, or `Bearish`
- `price_trend_quantitative`: Any numeric or percentage change mentioned (e.g. `+2.3%`, `-150 points`), or `Unknown`.
- `price_impact_qualitative`: Either `Positive` or `Negative`, indicating the expected directional impact on the NASDAQ index.

Below is a list containing only {column_title} extracted from factors of a single market report. Your task is to determine which of the candidate reports these factors most likely came from.
{feature_text}

The candidate reports are as follows:
{candidates}
\end{lstlisting}

\subsection{Article Summarisation Prompt}
\begin{lstlisting}[breaklines=true, basicstyle=\ttfamily]
Assume it is {date}. You will be given article URLs, headlines, metadata, and, where permitted by source terms, limited excerpts from news articles involving {topic}. Your task is to summarize the content in a way that is directly useful to a financial analyst writing a report on price movements of the NASDAQ-100 index. Include citations (source names and URLs). Do not reproduce passages verbatim and do not include content beyond what is allowed by the source's terms. Ensure you keep specific dates, financial metrics, price movements, and companies or tickers mentioned. Your summary should be succinct, accurate, and complete within these constraints.
{article_text}
\end{lstlisting}

\subsection{Report Generation Prompt}
\begin{lstlisting}[breaklines=true, basicstyle=\ttfamily]
Assume it is currently {date} and you are an expert trader and financial analyst. Below, you are given the historical NASDAQ-100 index closing price, combined with predicted closing price for this index one week into the future. You are also given the historic and predicted closing price for several stocks, as well as their earnings calendar for the next quarter. These stocks were selected based on importance for the time period.
Additionally, you are given some other important information about the US economy in the past quarter, including unemployment rates, initial jobless claims, inflation rates, GDP, federal funds rates and commodity prices. Your task is to generate a well written, insightful stock market report, providing reasoning for the predicted NASDAQ index price, based on the provided data. You should cite datapoints as much as possible, and include any units, or provide percentage changes. Use both the provided additional data, as well as external data. The report should be concise, informative and suitable for a professional audience.

You are provided with a list of dictionaries, each of which details the historic seasonality and trend of closing price for a given ticker, along with additional information about the ticker and data collection.
{ticker_dictionary}

Additionally, you may refer to the following external factors in your report. These factors have been extracted from news articles.
{external_news_summaries}
    
The following are both good and bad examples of market reports, along with explanations of what makes the report good or bad. You should follow the writing style and tone of good reports.
{examples}

Return your report in plain text, with no other formatting. This output is for research illustration only and is not investment advice. Where external information is referenced, include citations (source names and URLs) and avoid reproducing verbatim text.
\end{lstlisting}

\subsection{Contradiction Detection Prompt}
\begin{lstlisting}[breaklines=true, basicstyle=\ttfamily]
The following axioms are known to be true:
{axioms}

Given a claim, determine whether it is contradictory based only on the above statements. A statement is contradictory if it cannot be true given the axioms.

If the claim is contradictory, respond with 'False'. If not, respond with 'True'. 

Do not provide any additional information or reasoning.

For example:
Claim: The sky is blue. Label: True
Claim: The Earth is flat. Label: False

{claim} Label:
\end{lstlisting}

\subsection{Persuasiveness Judgement Prompt}
\begin{lstlisting}[breaklines=true, basicstyle=\ttfamily]
You are an expert stock market analyst. Your task is to compare given reports and determine which one is more convincing, if any. In this scenario, we define convincing as coherent, relevant, specific, references important data points and external factors, and most convinces you why it is the best explanation for the observed market price changes. There may be cases where neither report is significantly more convincing.

Return your answer as a single integer, indicating the number of the report you found most convincing, if any, and 0 if not. For instance, if you found report 1 most convincing, return 1. If you found no report to be more convincing than any other, return 0. 

Return only the integer without any additional text or explanation.
\end{lstlisting}

\section{Generated Report}
\label{app:generated_report}
A full generated report, generated for November 29, 2024, is provided below. 

\begin{lstlisting}[breaklines=true, basicstyle=\ttfamily]
November 29 2024 - 08:00 a.m. EST  
Market Commentary (Illustrative academic example; not investment advice; not affiliated with or endorsed by any organisation.)

Headline
The NASDAQ-100 continues to climb on a broad easing of macro-pressure.  Our model projects the index to close near 19,860 on 6 December and 19,636 on the week ending 8 December, implying a further 2.7 % gain from last week's settle at 19,127 (gradient ~ +195 pts/week).

Macro backdrop - still "Goldilocks"
-Employment: The national unemployment rate is flat at 3.9 % for a third straight month; the 20-week trend slope is fractionally negative (-0.002 ppt). Weekly initial jobless claims are falling sharply (trend -568 claims/week; last reading -15.6 % w/w or -39.8 k). 
-Inflation expectations: 10-yr breakevens eased 3 bp this week to 1.96 % and remain in a gentle 5-month up-trend of just +0.6 bp/week - low enough to keep real yields supportive.
-Growth: The Fed's high-frequency GDP proxy (BBK index) is drifting lower (-0.35 index-pt/week, latest -5.75 m/m), but the pace is orderly and consistent with a soft-landing narrative.  
-Rates: Effective fed-funds have edged down 19 bp over the past week (-3.9 % w/w) and the 5-month slope is negative (-4.1 bp/month), indicating markets are already pricing the first cut for Q2-25. 
-Commodities: WTI has slipped another 6.8 % w/w to USD 74.9/bbl (5-month slope -0.66 $/week), alleviating energy-led margin pressure.

Equity leadership - semiconductors still bid
Qualcomm (QCOM) exemplifies the bid for AI and handset exposure:  
-Daily series: Price up 4.9 % w/w to USD 242.46 on 5 Dec; 5-week gradient +0.68 $/day.
-Weekly series: Trending +3.0 $/week since early Nov; latest close USD 241.5 (+3 % w/w). 
-Earnings: Street is looking for FQ1-25 EPS USD 2.35 on 30 Jan; every 1 % revision to that number adds ~USD 1.30 to our DCF-derived fair value.  Guidance risk skews positive on lower silicon input costs and Apple modem volumes.

Sector implications of geopolitics
The freight-sensitive semiconductor supply chain remains the key tail-risk.  Rising U.S.-China/Taiwan tension could disrupt Asian lanes and raise spot container rates, but for now the market is discounting only a mild risk premium - chipmakers are up 6 % since the Nov 15 policy headline.  A material escalation would be index-negative yet relatively QCOM-neutral thanks to its diversified foundry strategy.

Outlook
With labour still tight, inflation expectations contained and crude in a down-channel, the Fed has room to signal an end to quantitative tightening at the 13 Dec FOMC.  That scenario underpins our year-end NASDAQ-100 target of 20,100 (~ +3.0 % from current).  Key data to watch:

-29 Nov PCE deflator - a print below 0.2 % m/m is the pivot cue.  
-6 Dec NFP - anything <150 k corroborates the claims trend.  
-OPEC+ meeting (postponed to 30 Nov) - deeper cuts could reverse the oil slide and cap equities around 20 k.

Risk skew: moderately positive.  A surprise re-acceleration in wages or a supply-shock in the Taiwan Strait would force a volatility spike; otherwise the path of least resistance remains upward into Q1-25.

Disclosure: The author holds no positions in the securities mentioned.
\end{lstlisting}
\section{\label{app:human_eval}Human Evaluation}

This appendix provides full details of the human evaluation conducted to qualitatively assess the quality of generated explanatory reports. The purpose of this evaluation was to complement automated metrics with expert human judgment on dimensions that are difficult to capture algorithmically, including stylistic appropriateness, interpretability, and practical usefulness for decision-making.

\subsection{Evaluator Selection}

Three evaluators were recruited for the study. All evaluators were screened to ensure domain familiarity and professional relevance. Specifically, evaluators were required to have read and used stock market reports in a professional or decision-support capacity within the month preceding the evaluation.

This selection criterion ensured that feedback reflected the expectations and interpretive standards of the intended audience for the generated reports.

\subsection{Evaluation Materials}

The evaluation focused on reports generated for the NASDAQ-100 case study. Due to time and resource constraints, three reports were randomly sampled from the generated corpus. Each report was presented independently and in randomized order to mitigate ordering effects and reduce anchoring bias.

Evaluators were not informed whether a report was machine-generated or human-authored.

\subsection{Evaluation Rubric}

Reports were assessed using a structured rubric comprising four components:
\begin{itemize}
    \item \textbf{Readability and Clarity}: clarity of expression, logical flow, and appropriateness of language for a professional audience.
    \item \textbf{Factual Consistency}: absence of internal contradictions and consistency with known domain facts.
    \item \textbf{Persuasiveness of Explanation}: effectiveness in linking evidence to conclusions and presenting a convincing narrative.
    \item \textbf{Overall Assessment}: perceived usefulness of the report for decision-making.
\end{itemize}

Each dimension was rated on a five-point Likert scale, with descriptive anchors provided to reduce ambiguity. Evaluators were additionally asked whether they would trust or use the explanation in practice.

The full evaluation rubric provided to evaluators is shown in Figure~\ref{fig:eval-rubrics}.

\subsection{Qualitative Feedback}

In addition to numeric ratings, evaluators were encouraged to provide free-form qualitative comments. These comments proved particularly valuable for identifying strengths and weaknesses not fully captured by scalar scores.

Recurring themes in the feedback included:
\begin{itemize}
    \item The importance of explicitly distinguishing between short-term and long-term effects of the same market driver.
    \item The heavy use of technical jargon, which may limit accessibility for non-specialist decision-makers.
    \item Excessive micro-level detail in some sections, particularly when daily gradients were reported for multiple assets.
    \item Strong risk framing, with clear links between drivers and potential catalysts, which evaluators viewed as a major strength.
\end{itemize}

These observations provided actionable guidance for refining prompt structure, evidence selection, and narrative balance in subsequent iterations of the generation pipeline.

\newpage
\begin{figure}[!ht]
    \centering
    \makebox[\textwidth][l]{\includegraphics[width=1.2\linewidth, trim={2cm 14cm 2cm 3cm}, clip]{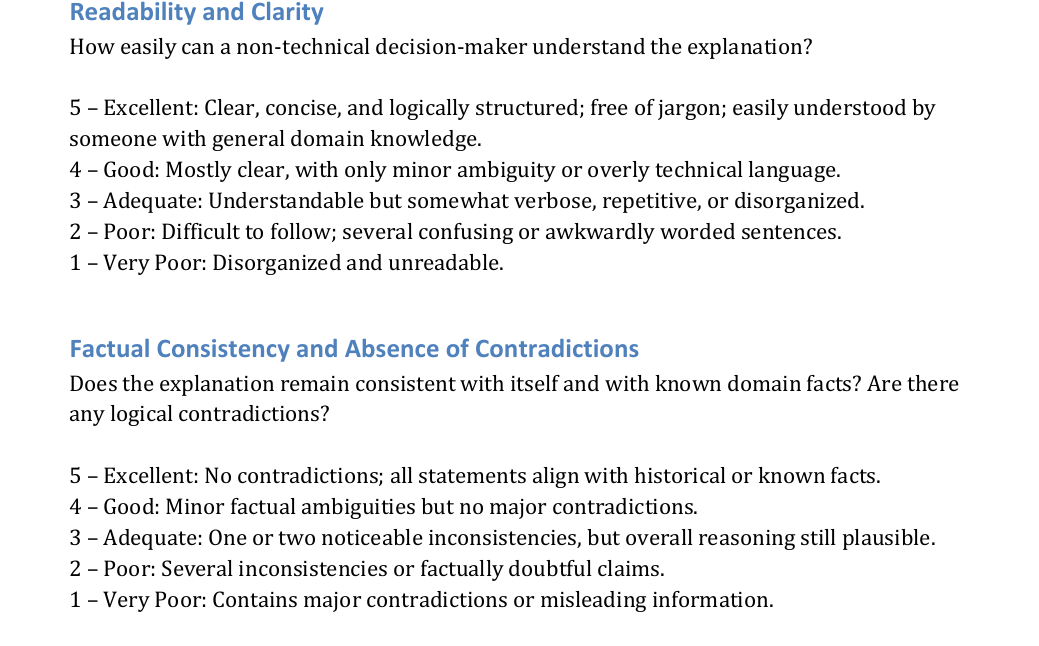}}
\end{figure}
\begin{figure}[!ht]
    \centering
    \makebox[\textwidth][l]{\includegraphics[width=1.2\linewidth, trim={2cm 4cm 2cm 14cm}, clip]{figures/Evaluation_Framework.pdf}}
    \caption{Full evaluation framework provided to human evaluators.}
    \label{fig:eval-rubrics}
\end{figure}

\end{document}